\PassOptionsToPackage{table}{xcolor}
\PassOptionsToPackage{pagebackref,breaklinks=true,colorlinks,citecolor=blue,urlcolor=blue,linkcolor=blue,bookmarks=false}{hyperref}
\PassOptionsToPackage{noabbrev,nameinlink,capitalize}{cleveref}

\documentclass[onecolumn, numbers]{april_aigc}

\usepackage[utf8]{inputenc} 
\usepackage{url}            
\usepackage{amsfonts}       
\usepackage{amssymb}        
\usepackage{amsmath}        
\usepackage{nicefrac}       
\usepackage{microtype}      
\usepackage{xspace}         
\usepackage{fix-cm}
\usepackage[T1]{fontenc}
\usepackage{ragged2e}
\usepackage{booktabs}       
\usepackage{float}          

\usepackage{wrapfig}        
\usepackage{multicol}       
\usepackage{multirow}       
\usepackage{makecell}       
\usepackage{tabularx}       
\usepackage{adjustbox}      
\usepackage{colortbl}       
\usepackage{pifont}         
\usepackage{enumitem}
\usepackage[normalem]{ulem}

\usepackage{xcolor}
\definecolor{linkcolor}{named}{aprilblue}
\definecolor{urlcolor}{RGB}{255,105,180}
\definecolor{citecolor}{RGB}{66,168,235}
\definecolor{lightgray}{rgb}{0.8, 0.8, 0.8}
\definecolor{darkgreen}{rgb}{0.00, 0.81, 0.78}

\definecolor{gray_tab}{RGB}{220, 220, 220}
\definecolor{blue_tab}{RGB}{227, 240, 251}
\definecolor{oran_tab}{RGB}{252, 242, 237}
\definecolor{whit_tab}{RGB}{255, 255, 255}
\definecolor{green_code}{RGB}{55, 126, 34}

\usepackage{algorithm}
\usepackage{algorithmic}
\usepackage{listings}
\usepackage{etoolbox}

\makeatletter
\AfterEndEnvironment{algorithm}{\let\@algcomment\relax}
\AtEndEnvironment{algorithm}{\kern2pt\hrule\relax\vskip3pt\@algcomment}
\let\@algcomment\relax
\newcommand\algcomment[1]{\def\@algcomment{\footnotesize#1}}
\renewcommand\fs@ruled{\def\@fs@cfont{\bfseries}\let\@fs@capt\floatc@ruled
  \def\@fs@pre{\hrule height.8pt depth0pt \kern2pt}%
  \def\@fs@post{}%
  \def\@fs@mid{\kern2pt\hrule\kern2pt}%
  \let\@fs@iftopcapt\iftrue}
\makeatother

\usepackage[pagebackref,breaklinks=true,colorlinks,citecolor=blue,urlcolor=blue,linkcolor=blue,bookmarks=false]{hyperref}
\AtEndPreamble{
    \usepackage[capitalize]{cleveref}
    \crefname{section}{Sec.}{Secs.}
    \Crefname{section}{Section}{Sections}
    \crefname{table}{Tab.}{Tabs.}
    \Crefname{table}{Table}{Tables}
    \crefname{equation}{Eq.}{Eqs.}
    \Crefname{equation}{Equation}{Equations}
    \crefname{figure}{Fig.}{Figs.}
    \Crefname{figure}{Figure}{Figures}
}
\hypersetup{colorlinks=true,linkcolor=linkcolor,urlcolor=urlcolor,citecolor=citecolor}

\usepackage{caption}
\DeclareCaptionFormat{custom}{{\color{aprilblue}\sffamily\textbf{#1 #2}} #3}
\titleformat*{\section}{\color{aprilblue}\Large\sffamily\bfseries}
\titleformat*{\subsection}{\color{aprilblue}\large\sffamily\bfseries}
\titleformat*{\subsubsection}{\color{aprilblue}\normalsize\sffamily\bfseries}

\usepackage{fancyhdr}
\newif\ifshowlogo
\showlogotrue   
\newcommand{\insertlogo}{%
  \ifshowlogo
    \IfFileExists{assets/april_logo1.png}%
    {\includegraphics[height=0.68cm]{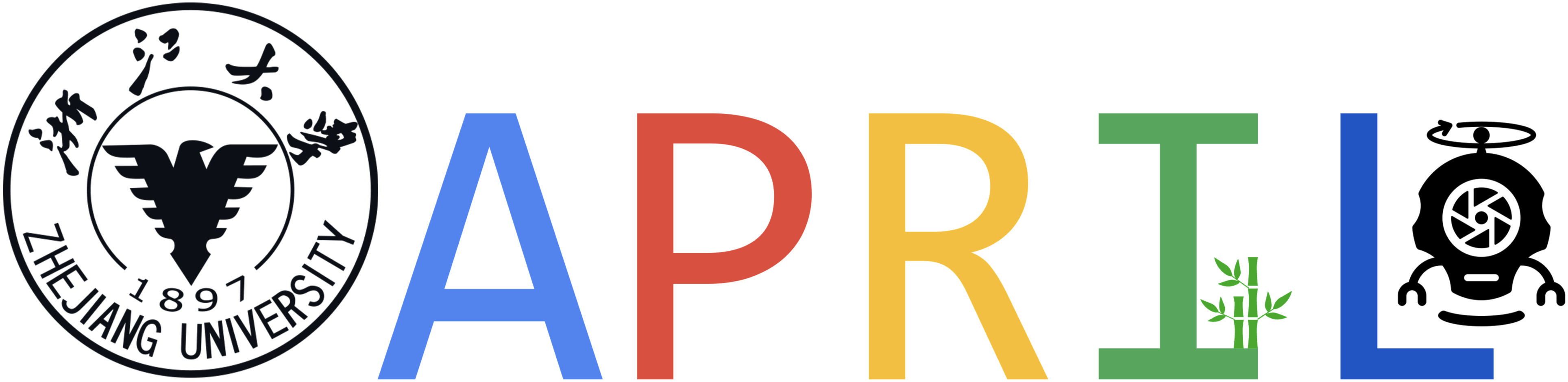}}
    {}%
  \fi
}
\newif\ifshowtoc
\showtocfalse

\renewcommand{\title}[1]{\def\titlelist{{\fontsize{20pt}{28pt}\selectfont\sffamily\bfseries #1}}}
\title{TAT-Det: A Dataset for Testicular Appendage Torsion Detection and Segmentation in Ultrasound Images}

\author[1,\star]{Ying Jiang}

\author[2,\star]{Juntao Jiang}

\author[3]{Linxuan Fan}

\author[1]{Lingling Li}

\author[4]{Yaqi Wang}

\author[1]{Xiuzhen Yang}

\author[2]{Yong Liu}

\author[2,\dagger]{Jiangning Zhang}

\author[1,\dagger]{Jingjing Ye}

\affiliation[1]{Department of Ultrasound, Children's Hospital, Zhejiang University School of Medicine, National Clinical Research Center for Children and Adolescents' Health and Diseases, Hangzhou, Zhejiang, China}

\affiliation[2]{College of Control Science and Engineering, Zhejiang University, Hangzhou, Zhejiang, China}

\affiliation[3]{Data Science Institute, Vanderbilt University, Nashville, Tennessee, USA}

\affiliation[4]{Innovation Center for Electronic Design Automation Technology, Hangzhou Dianzi University, Hangzhou, Zhejiang, China}

\contribution[\star]{Equal contribution}

\contribution[\ddagger]{Corresponding author}

\abstract{
\textbf{Background}: Testicular appendage torsion (TAT) is the leading cause of acute scrotal pain in prepubertal children. Yet, its diagnosis remains challenging due to nonspecific clinical manifestations and the absence of standardized datasets for pediatric acute scrotal emergencies. 

\textbf{Results}: We present TAT-Det, an expert-annotated ultrasound dataset specifically designed for TAT. TAT-Det comprises 2,600 ultrasound images from 1,203 patients collected at the Children's Hospital, Zhejiang University School of Medicine. Each TAT-positive image is accompanied by lesion-level bounding boxes for detection and pixel-level segmentation masks for fine-grained delineation. Annotations were produced through a multi-stage workflow involving independent annotation by board-certified pediatric sonographers, consensus review, and refinement. Baseline benchmarks for object detection and semantic segmentation using representative deep learning models are further provided, with YOLO11m achieving the best detection performance and RWKV-UNet the best segmentation performance. 

\textbf{Conclusions}: TAT-Det addresses the lack of publicly available resources for pediatric emergency ultrasound and establishes a foundation for developing and evaluating artificial intelligence methods for automated TAT analysis.
}

\coverdate{\today}

\begin{document}

\maketitle
\thispagestyle{plain}

\ifshowtoc
    \clearpage
    \setcounter{tocdepth}{2} 
    
    \tableofcontents
    \vspace{1cm} 


    \clearpage
\fi

\section{Introduction}
\label{sec:introduction}

Testicular appendage torsion (TAT) is the leading cause of acute scrotum in prepubertal boys, accounting for 31\%--67\% of all pediatric acute scrotal emergencies~\cite{makela200719,kim2018clinical,deeg2014doppler}. Anatomically situated at the superior pole of the testis, the appendix testis is a vestigial remnant of the paramesonephric (M\"ullerian) duct, typically manifesting as a pedunculated, oval structure measuring up to 7--8 mm in length~\cite{laimer2022multiparametric}. This pedunculated morphology predisposes the appendage to axial torsion, precipitating vascular compromise and sudden-onset scrotal pain. Although TAT is inherently benign and conservatively managed, its clinical presentation closely mimics that of testicular torsion—a true surgical emergency requiring immediate intervention to salvage the testis. Consequently, the rapid and definitive identification of TAT is of paramount clinical importance to guide appropriate triage and obviate unnecessary surgical exploration.

Clinically, differentiating TAT from other etiologies of acute scrotum remains formidable due to the substantial overlap in nonspecific symptoms, such as acute pain, erythema, and swelling~\cite{varga2007acute,langan2022scrotal}. While the pathognomonic ``blue dot sign'' is highly specific for TAT, it is clinically discernible in only a minority of patients~\cite{langan2022scrotal}. Reliance on physical examination alone is therefore fraught with diagnostic uncertainty, yielding misdiagnosis rates as high as 45\% among general practitioners and 7\%--10\% even among seasoned pediatric urologists~\cite{baldisserotto2005color}. These diagnostic pitfalls underscore the critical need for objective, imaging-based evaluation in pediatric emergency settings.

Scrotal ultrasonography (US), coupled with color Doppler, is universally endorsed as the first-line imaging modality for acute scrotum assessment, owing to its non-ionizing nature, real-time capability, and high accessibility~\cite{sung2012sonography,gunther2012acute}. Sonographically, a torsed appendage typically presents on grayscale imaging as a discrete, avascular nodule wedged between the testis and epididymis. However, its acoustic signature is highly dynamic and stage-dependent: it appears enlarged and hypoechoic within the first 24 hours of onset (mean diameter: $8.0 \pm 3.1$ mm), before progressively undergoing atrophy and becoming hyperechoic (mean diameter: $3.7 \pm 0.9$ mm)~\cite{deeg2021differential}. The minute scale of the lesion, its evolving echogenicity, and its intimate proximity to adjacent anatomical structures render consistent sonographic identification notoriously difficult. Furthermore, US interpretation is inherently operator-dependent and subject to substantial inter-observer variability, particularly in resource-limited or general emergency settings lacking dedicated pediatric sonographers~\cite{ciftci2004clinical}. These clinical and technical bottlenecks strongly motivate the development of automated, computer-aided diagnostic (CAD) tools to standardize TAT localization and assist clinicians in high-stakes decision-making.

From a computational perspective, the precise spatial localization of the torsed appendage is the cornerstone of a reliable CAD system. Unlike black-box, image-level classification, object localization provides explicit anatomical grounding and clinically interpretable outputs, which are indispensable for building trust in emergency triage scenarios. Specifically, bounding-box representations offer a pragmatic and robust paradigm for capturing minute lesions in US imagery, where acoustic shadowing, speckle noise, and ill-defined boundaries frequently confound pixel-level delineation. Establishing a standardized benchmark for both coarse localization (detection) and fine-grained delineation (segmentation) is therefore vital for advancing algorithmic development in this domain.

Despite the rapid proliferation of medical imaging datasets, publicly accessible benchmarks for pediatric scrotal emergencies remain virtually nonexistent. In recent years, open-access US datasets have catalyzed breakthroughs in deep learning across various anatomical sites, including the breast (e.g., BUSI~\cite{al2020dataset}, BUS-BRA~\cite{gomez2024bus}, UDIAT~\cite{yap2017automated}), thyroid (e.g., DDTI~\cite{pedraza2015open}, TN3K~\cite{gong2021multi}), heart (e.g., CAMUS~\cite{leclerc2019deep}, EchoNet-Dynamic~\cite{ouyang2020video}), and fetus (e.g., HC18~\cite{van2018automated}). While these resources have standardized the evaluation of classification, segmentation, and functional assessment algorithms, pediatric acute scrotum has been largely overlooked. The absence of a dedicated, publicly available dataset for TAT constitutes a critical bottleneck, impeding the development and reproducible benchmarking of AI systems tailored to pediatric emergency ultrasound.

To bridge this gap, we introduce \textbf{TAT-Det}, the first publicly available, expert-annotated ultrasound dataset specifically curated for testicular appendage torsion—and, to our knowledge, the first open-access resource dedicated to pediatric acute scrotal emergencies. TAT-Det comprises 2,600 high-resolution US images retrospectively collected from the Children's Hospital, Zhejiang University School of Medicine, spanning a three-year period from September 2022 to August 2025. Moving beyond the standard classification or isolated segmentation paradigms of existing US benchmarks, TAT-Det formulates lesion localization as its primary objective while simultaneously supporting fine-grained morphological analysis. Every image in the cohort is meticulously annotated with both bounding boxes and pixel-level segmentation masks. To ensure gold-standard reliability, all annotations were executed and cross-verified under a rigorous consensus protocol by board-certified pediatric sonographers with over ten years of specialized clinical experience.

In summary, TAT-Det establishes a foundational benchmark for developing and evaluating AI algorithms in pediatric scrotal ultrasonography. By opening a clinically vital yet computationally underexplored domain, this dataset aims to: (1) catalyze the development of robust algorithms for small-lesion detection and segmentation amidst complex acoustic artifacts; (2) facilitate reproducible, fair comparisons across state-of-the-art architectures; and (3) ultimately accelerate the translation of AI-assisted diagnostic tools into pediatric emergency care.

\paragraph{Key points}
\begin{itemize}
\item TAT-Det is the first publicly available ultrasound dataset dedicated to testicular appendage torsion, comprising 2,600 ultrasound images from 1,203 pediatric patients.
\item Each TAT-positive image carries expert-annotated lesion-level bounding boxes and pixel-level segmentation masks produced through a consensus-based multi-stage annotation workflow.
\item Baseline benchmarks across seven YOLO generations and nine segmentation architectures establish a standardized evaluation foundation for pediatric acute scrotal ultrasound analysis.
\end{itemize}
\section{Related Work}
\label{sec:related_work}

\paragraph{Comparison with existing ultrasound datasets}
Table~\ref{tab:ultrasound_dataset_comparison} positions TAT-Det relative to
several widely used public ultrasound datasets spanning breast, thyroid, cardiac,
and fetal imaging. Three distinctions emerge.

\textbf{Anatomical coverage.} Existing public benchmarks concentrate on
well-established screening targets: breast lesions
(BUSI~\cite{al2020dataset}, BUS-BRA~\cite{gomez2024bus},
UDIAT~\cite{yap2017automated}), thyroid nodules
(DDTI~\cite{pedraza2015open}, TN3K~\cite{gong2021multi}), and cardiac or
fetal structures (
EchoNet-Dynamic~\cite{ouyang2020video}, HC18~\cite{van2018automated}).
Testicular appendage pathology, despite being the most common cause of acute
scrotal pain in children, has no
dedicated public dataset to date. TAT-Det closes this gap and extends the
scope of ultrasound-based computer-aided diagnosis to an under-served
urological emergency.

\textbf{Multi-task annotation support.} Beyond the masks, contours, ellipses, and
image-level labels offered by existing datasets, TAT-Det provides two spatially
aligned levels of annotation for every TAT-positive image, with empty labels and all-zero masks for
lesion-free images. This allows detection and segmentation to be trained and
evaluated on identical images and patient-level splits, and supports joint or
localization-aware multi-task learning.

\textbf{Scale and task coupling.}  As summarized in Table~\ref{tab:ultrasound_dataset_comparison}, TAT-Det provides 2,600 ultrasound images of testicular appendages, placing it among the larger publicly available datasets for lesion- or structure-oriented ultrasound analysis. Its scale exceeds that of BUSI~\cite{al2020dataset} (780), BUS-BRA~\cite{gomez2024bus} (1,875), UDIAT~\cite{yap2017automated} (163), DDTI~\cite{pedraza2015open} (637), and HC18~\cite{van2018automated} (1,334), and approaches that of TN3K~\cite{gong2021multi} (3,493), although these datasets primarily focus on breast, thyroid, or fetal imaging. In contrast, TAT-Det targets the substantially less studied anatomy of the testicular appendage in pediatric scrotal ultrasound, addressing a gap that is not represented by the existing benchmarks. Importantly, TAT-Det is not limited to a single annotation modality: each TAT-positive image is accompanied by lesion-level bounding boxes for localization and pixel-level masks for fine-grained delineation. This combination enables the dataset to support both object detection and segmentation, while also facilitating studies of localization-aware segmentation and other downstream computer-aided analysis tasks. Compared with datasets that provide only contours, masks, or image-level diagnostic labels, TAT-Det therefore couples a relatively substantial dataset scale with complementary spatial annotations in an anatomically underrepresented clinical domain.

\begin{table*}[!ht]
\centering
\small
\renewcommand{\arraystretch}{0.9}
\caption{Comparison between TAT-Det and representative publicly available ultrasound datasets.}
\label{tab:ultrasound_dataset_comparison}
\resizebox{\textwidth}{!}{%
\begin{tabular}{l l c p{3.4cm} p{3.3cm}}
\toprule
\textbf{Dataset} & \textbf{Organ / Anatomy} & \textbf{Images} & \textbf{Annotation Type} & \textbf{Task Type} \\
\midrule
BUSI~\cite{al2020dataset}
& Breast lesion
& 780
& Pixel-level masks; lesion category labels
& Segmentation; classification \\

BUS-BRA~\cite{gomez2024bus}
& Breast lesion
& 1,875
& Pixel-level masks; BI-RADS labels
& Segmentation; BI-RADS grading \\

UDIAT~\cite{yap2017automated}
& Breast lesion
& 163
& Pixel-level masks; lesion labels
& Segmentation; classification \\

DDTI~\cite{pedraza2015open}
& Thyroid nodule
& 637
& Nodule contours
& Segmentation \\

TN3K~\cite{gong2021multi}
& Thyroid nodule
& 3,493
& Pixel-level masks
& Segmentation \\

EchoNet-Dynamic~\cite{ouyang2020video}
& Heart (echocardiography)
& 10,030 videos
& Ventricle tracings; ejection-fraction values
& Video segmentation; cardiac function regression \\

HC18~\cite{van2018automated}
& Fetal head
& 1,334
& Head-circumference ellipse
& Segmentation; biometric measurement \\

\textbf{TAT-Det}
& \textbf{Testicular appendage}
& \textbf{2,600}
& \textbf{Bounding boxes + pixel-level masks}
& \textbf{Detection + segmentation} \\

\bottomrule
\end{tabular}%
}
\end{table*}
\section{Data Description}
\label{sec:data}

\begin{figure}[H]
    \centering
    \includegraphics[width=0.8\linewidth]{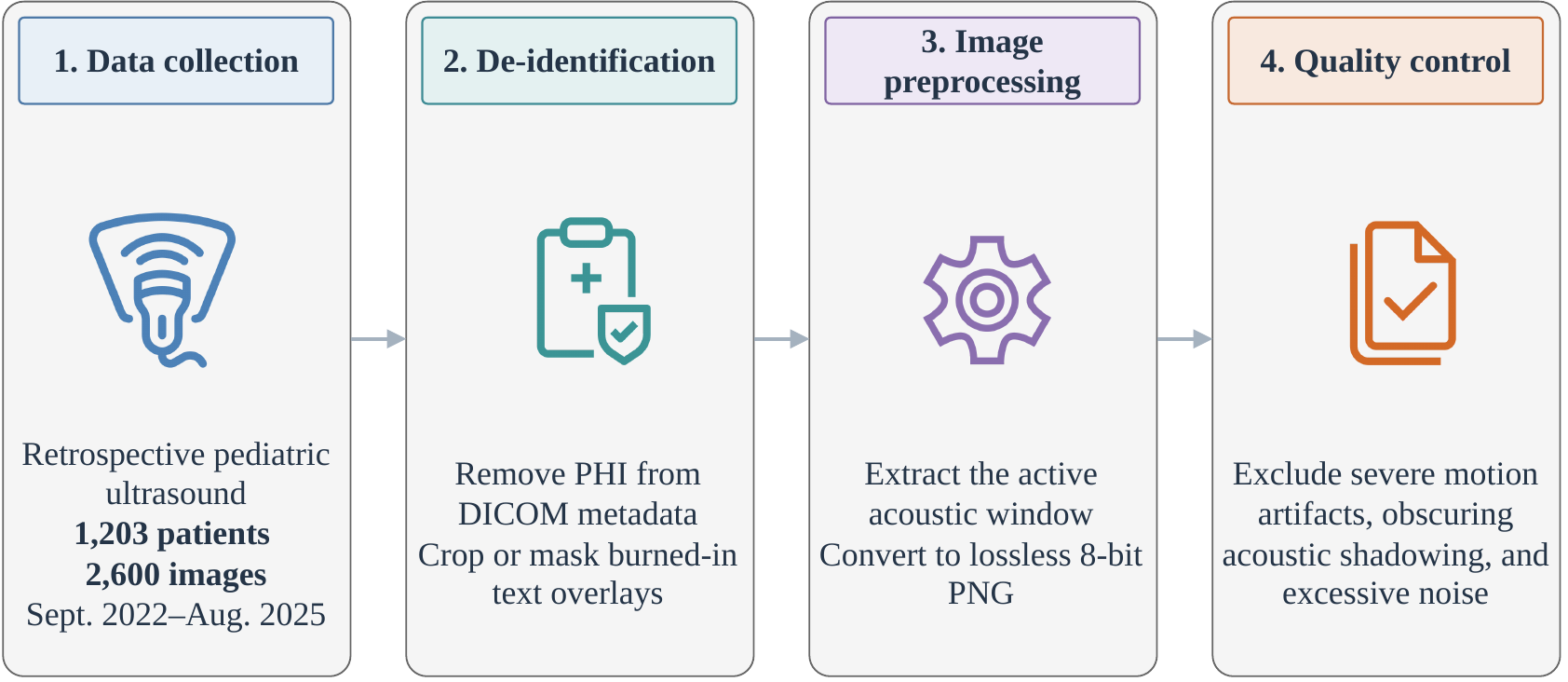}
   \caption{TAT-Det Data Collection and Preprocessing Workflow: Retrospectively collected pediatric ultrasound data are de-identified, converted into lossless 8-bit PNG images through active acoustic-window extraction, and quality-screened to exclude severe artifacts and noise.}
    \label{fig:tat_dataset_workflow1}
\end{figure}

\begin{figure*}[!ht]
    \centering
    \includegraphics[width=0.88\linewidth,height=0.45\textheight,keepaspectratio]{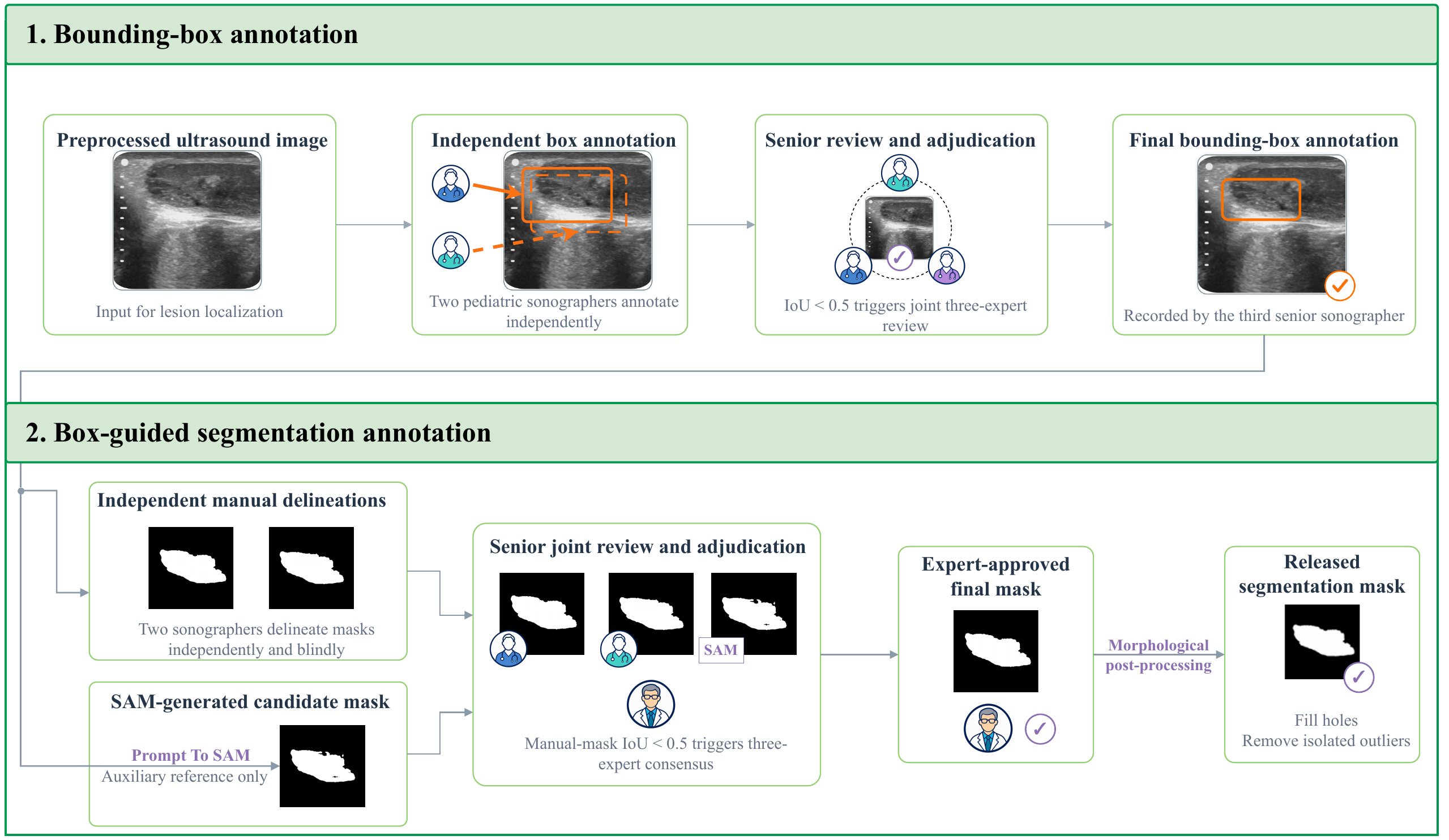}
   \caption{TAT-Det Annotation Workflow: Independent bounding-box annotation and expert consensus are followed by independent mask annotation, senior review, and morphological post-processing.}
    \label{fig:tat_dataset_workflow2}
\end{figure*}

The TAT-Det dataset was constructed following a systematic workflow comprising clinical ultrasound data collection, image preprocessing and quality control, expert annotation, annotation refinement, and patient-level dataset partitioning. These stages collectively transformed raw clinical examinations into a standardized benchmark with both lesion localization and pixel-level segmentation supervision, while ensuring data quality, annotation reliability, and patient-level independence between dataset splits. The overall data acquisition and preprocessing procedures, including ultrasound examination collection, DICOM-based preprocessing, and image quality control, are illustrated in Fig.~\ref{fig:tat_dataset_workflow1}. The subsequent annotation workflow, including independent expert annotation, expert consensus, and annotation refinement, is presented in Fig.~\ref{fig:tat_dataset_workflow2}.

\begin{figure*}[!ht]
\centering
\includegraphics[width=\textwidth,height=0.45\textheight,keepaspectratio]{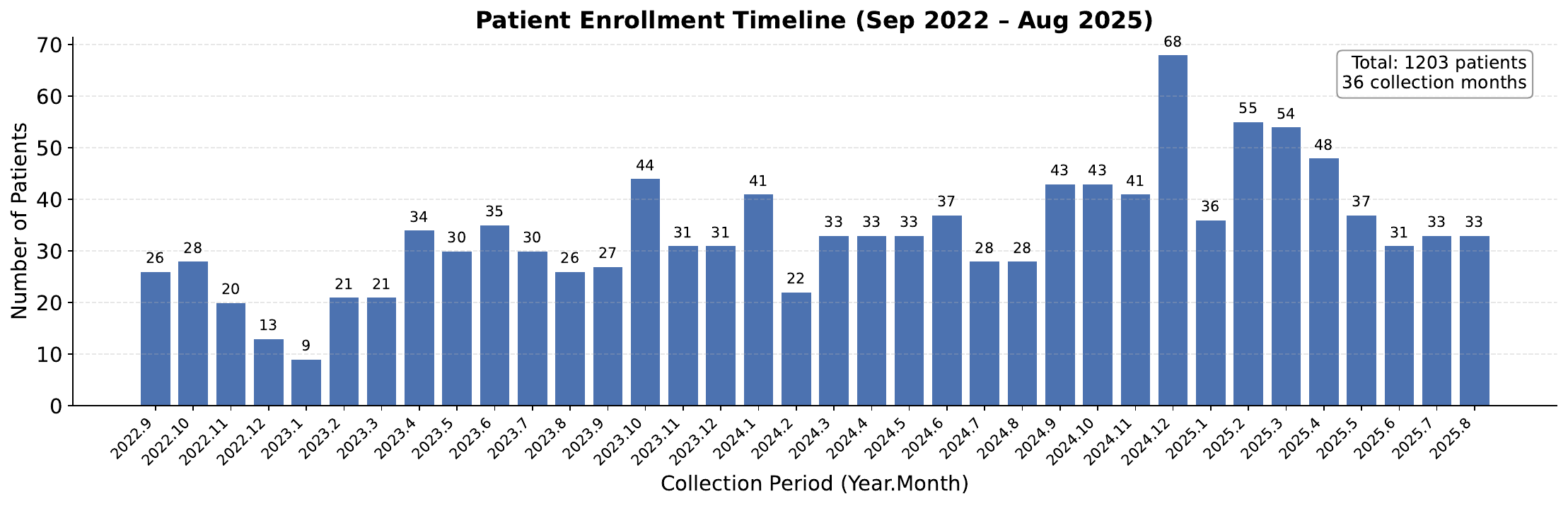}
\caption{Patient enrollment timeline from September 2022 to August 2025. A total of 1,203 patients were enrolled across 36 collection months.}
\label{fig:timeline}
\end{figure*}

\paragraph{Data collection} The dataset comprises ultrasound images retrospectively collected from pediatric patients who underwent scrotal ultrasound examinations for acute scrotal pain at the Children's Hospital, Zhejiang University School of Medicine, between September 2022 and August 2025. The patient enrollment timeline can be seen in Fig.~\ref{fig:timeline}. The initial cohort includes 2,600 ultrasound images from 1,203 patients. The median age is 9.8 years, with the interquartile range spanning 7.9--11.2 years. The most common age is 10 years ($n = 213$, 17.7\%). Detailed age statistics are summarized in Table~\ref{tab:age_stats}. The images are acquired using multiple ultrasound platforms from three manufacturers (GE Healthcare, Philips, and Mindray), with detailed equipment information summarized in Table~\ref{tab:ultrasound_machines}. This diversity in imaging platforms introduces realistic variations in image quality, acquisition characteristics, and visual appearance, improving the dataset's robustness and its potential to support the development and evaluation of models with better cross-device generalizability. Patient-level positive and negative labels were determined according to the final clinical diagnoses documented in the medical records, which served as the reference standard. For TAT-positive cases, the diagnosis was further supported by characteristic sonographic findings and explicit documentation in the official ultrasound reports. For each patient, two to three frames were selected from the initial examination by a sonographer with more than 10 years of experience. Frames were selected to represent standard diagnostic views while preserving a range of lesion appearances. Images with severe artifacts, incomplete lesion visualization, or duplicated acquisition views were excluded.

\begin{table}[H]
\caption{Age distribution of pediatric patients in the TAT-Det dataset (unit: years).}
\label{tab:age_stats}
\centering
\footnotesize
\setlength{\tabcolsep}{3pt}
\begin{tabular*}{\textwidth}{@{\extracolsep{\fill}} c c c c c c c c@{}}
\toprule
\textbf{Min} & \textbf{Max} & \textbf{Mean} & \textbf{SD} & \textbf{Median} & \textbf{IQR} & \textbf{Q1} & \textbf{Q3} \\
\midrule
0.2 & 17.2 & 9.3 & 2.7 & 9.8 & 3.3 & 7.9 & 11.2 \\
\bottomrule
\end{tabular*}
\end{table}
\begin{table}[H]
\caption{Ultrasound machines used in the TAT-Det dataset.}
\label{tab:ultrasound_machines}
\centering
\begin{tabular*}{\textwidth}{@{\extracolsep{\fill}} l l r @{}}
\toprule
\textbf{Manufacturer} & \textbf{Model} & \textbf{Patients} \\
\midrule
GE Healthcare & GE LOGIQ E20 & 518 \\
GE Healthcare & GE LOGIQ E9  & 325 \\
Philips       & Philips      & 232 \\
Mindray       & Resona R9T   &  69 \\
Mindray       & Resona A20T  &  17 \\
Mindray       & Resona 7S    &   1 \\
\midrule
Not recorded  & ---          &  41 \\
\midrule
\multicolumn{2}{l}{\textbf{Total}} & \textbf{1{,}203} \\
\bottomrule
\end{tabular*}
\end{table}

\paragraph{Data processing and anonymization} To construct a standardized and deep learning-ready dataset, a rigorous data pipeline was established, comprising de-identification, format conversion, and image preprocessing. To strictly protect patient privacy in compliance with the Declaration of Helsinki, all raw DICOM (Digital Imaging and Communications in Medicine) files underwent a comprehensive, two-stage de-identification process. All Protected Health Information (PHI) embedded in the DICOM metadata (e.g., patient name, medical record number, age, gender, and examination date) was fully anonymized. Additionally, any burned-in text overlays containing sensitive information on the ultrasound screen (such as hospital names, system parameters, and operator initials) were programmatically cropped or masked using a custom script. After automated de-identification, all processed images underwent independent cross-checking by two reviewers to verify the completeness of anonymization and confirm that no residual personally identifiable information remained. Besides, the processed DICOM files were converted into high-quality, lossless 8-bit PNG (Portable Network Graphics) format to facilitate algorithmic training. During this conversion, only the active acoustic window containing the scrotal anatomy was extracted, effectively discarding peripheral user-interface borders, color bars, and measurement scales that could otherwise introduce shortcut learning for neural networks. Finally, a strict quality control screen was performed: images with severe motion artifacts, acoustic shadowing that obscured the target pathology, or excessive noise were excluded.

\paragraph{Bounding box annotation} To establish a highly reliable benchmark for lesion localization, all bounding boxes targeting the torsed appendages and adjacent structures were annotated using the open-source tool \textit{LabelMe}~\cite{Wada_Labelme_Image_Polygonal}. Two pediatric sonographers with more than 5 years of experience independently annotated each case. The annotations were then reviewed by a third pediatric sonographer with 10 years of experience, who adjudicated the two independent results and provided the final bounding box annotation. Inter-rater agreement between two sonographers for primary bounding box annotations can be seen in Table~\ref{tab:inter_rater_iou}. For cases in which the bounding boxes produced by the two primary annotators exhibited an intersection-over-union (IoU) below 0.5, all three sonographers jointly review and discussed the case to reach a clinical consensus. Following this consensus discussion, the third senior sonographer determined and recorded the final annotation, thereby ensuring the clinical accuracy and consistency of the detection labels.

\paragraph{Segmentation mask generation} For the auxiliary segmentation masks, we
adopted a multi-stage annotation pipeline to ensure accurate delineation and
topological consistency despite the fuzzy boundaries and acoustic artifacts
inherent in ultrasound images. Mask annotation was performed on top of the
finalized detection labels: for each lesion, the consensus bounding box obtained in
the detection annotation stage was provided to the annotators as a reference
region, so that the structure to be delineated was unambiguously specified and the
resulting masks remained spatially consistent with the released boxes. Two
pediatric sonographers then independently delineated the torsed appendage at the
pixel level within the given bounding boxes using our custom-developed interactive
medical annotation software, without access to each other's annotations. In
parallel, the same consensus bounding boxes were used as prompts to the Segment
Anything Model (SAM)~\cite{kirillov2023segment} to generate a candidate mask for
each lesion. A third, more experienced pediatric sonographer then jointly reviewed
the two independent manual masks together with the SAM-generated candidate and
determined the final segmentation mask; the SAM output served only as an auxiliary
reference and was never adopted without expert verification. For cases in which the
two independent annotations exhibited an IoU below 0.5, all three sonographers
jointly inspected the ultrasound image and discussed the anatomical boundaries to
reach a consensus, after which the senior sonographer produced the final mask based
on the consensus decision. Inter-rater agreement between the two independent
delineations is reported in Table~\ref{tab:inter_rater_iou}. Finally, the finalized
masks underwent a dedicated morphological post-processing step to remove isolated
outlier pixels and fill missing regions within connected components, thereby
eliminating spurious disconnected regions and ensuring the spatial continuity and
structural integrity of the released segmentation masks.

\paragraph{Inter-rater agreement} As shown in Table~\ref{tab:inter_rater_iou}, the two raters demonstrated strong overall agreement for both lesion localization and delineation. For bounding-box annotations, the mean IoU reached 0.803, with a median of 0.841 and only 69 cases (4.0\%) exhibiting an IoU below 0.5. Pixel-level annotations showed a higher mean Mask mIoU of 0.838 and a median of 0.938, although 160 cases (9.3\%) had an mIoU below 0.5. The relatively high median values indicate that the two sonographers generally produced closely aligned annotations, while the larger standard deviations, particularly for mask annotations, reflect a small subset of challenging cases with greater inter-rater variability. These cases may arise from ambiguous lesion boundaries, heterogeneous sonographic appearance, or limited visibility of the appendage. Overall, the high agreement across both bounding-box and pixel-level annotations supports the reproducibility and reliability of the primary annotations used in TAT-Det.

\begin{table}[H]
\caption{Inter-rater agreement between two sonographers for primary annotations on 1,725 images with TAT.}
\label{tab:inter_rater_iou}
\centering
\footnotesize
\setlength{\tabcolsep}{2pt} 
\begin{tabular*}{\textwidth}{@{\extracolsep{\fill}} l c c c c c }
\toprule
\textbf{Metric (\%)} & \textbf{N} & \textbf{Mean} & \textbf{Median} & \textbf{Std} & \textbf{IoU$<$50\%, n (\%)} \\
\midrule
BBox IoU  & 1{,}725 & 80.3 & 84.1 & 15.0 & 69 (4.0) \\
Mask IoU  & 1{,}725 & 83.8 & 93.8 & 23.9 & 160 (9.3) \\
\bottomrule
\end{tabular*}
\end{table}

\paragraph{Dataset split}
The dataset was partitioned at the patient level, resulting in image-level counts of 2{,}021 training, 289 validation, and 290 test images. Both positive (lesion-bearing) and negative (lesion-free) samples were included in every split to ensure that the model learned to detect lesions while also suppressing false positives on normal tissue. Patient-level splitting was enforced to prevent data leakage: all images from the same patient were confined to a single split, ensuring that no patient appeared in more than one subset. This patient-independent partitioning strategy guaranteed that model performance on the validation and test sets reflected generalization to unseen patients rather than memorization of individual-specific patterns.

\begin{figure}[H]
    \centering
    \includegraphics[width=1\linewidth,height=0.45\textheight,keepaspectratio]{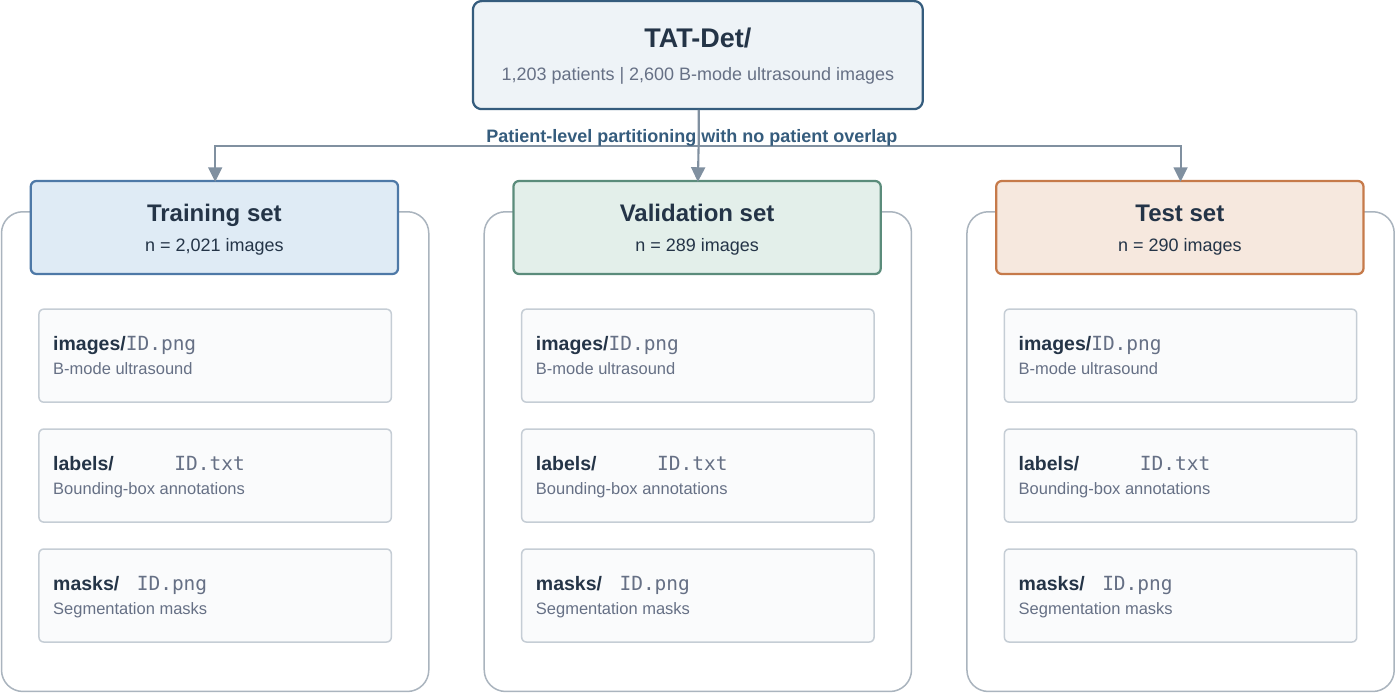}
   \caption{Hierarchical directory organization of the TAT-Det dataset. The identical sub-directory configurations across all three splits ensure compatibility with diverse deep learning pipelines for joint detection and segmentation.}
    \label{fig:dataset-structure}
\end{figure}

\paragraph{File organization and formats}
The TAT-Det dataset is available upon request under the controlled-access terms described in the Data availability section (the data application form can be downloaded from \url{https://raw.githubusercontent.com/juntaoJianggavin/TAT-Det/main/TAT-Det_Data_Application_Form.pdf}), and is structured to support both object detection and semantic segmentation workflows. The repository is organized into a standardized, hierarchical directory structure designed for seamless integration with modern deep learning frameworks. Each of the three subset directories (\texttt{train/}, \texttt{val/}, and \texttt{test/}) maintains an identical internal structure, containing raw ultrasound frames (\texttt{images/}), standard object detection bounding box annotations (\texttt{labels/}), and auxiliary pixel-level segmentation masks (\texttt{masks/}). This unified layout allows for seamless integration with multi-task learning paradigms. The hierarchical directory organization of the TAT-Det dataset is illustrated in Fig.~\ref{fig:dataset-structure}.

To ensure maximum compatibility and prevent diagnostic information loss, we adhere to the following file specifications across different modalities:

\begin{itemize}
    \item \textbf{Source Images (\texttt{images/}):} All ultrasound frames are stored in \texttt{.png} format with 8-bit depth. To preserve the original acoustic details, no lossy compression (such as JPEG) is applied. Resolution varies based on the acquisition preset of the ultrasound system.

    \item \textbf{Detection Annotations (\texttt{labels/}):} Bounding box labels are provided in the standard \textbf{YOLO text format}. For each image, a corresponding \texttt{.txt} file contains one or more lines (representing annotated lesions). Each line contains five space-separated values:

    $\langle\mathrm{class\_id}\rangle \quad \langle x_{\mathrm{center}}\rangle \quad \langle y_{\mathrm{center}}\rangle \quad \langle\mathrm{width}\rangle \quad \langle\mathrm{height}\rangle$,
    where the class ID for a torsed appendix is set to \texttt{0}, and all spatial coordinates are normalized to the range $[0, 1]$ relative to the width and height of the image.

    \item \textbf{Segmentation Masks (\texttt{masks/}):} The auxiliary segmentation masks are stored as single-channel 8-bit grayscale PNG images that match the exact spatial dimensions of their corresponding source images. Pixels belonging to the background are represented by value \texttt{0}, while pixels representing the torsed appendage lesion are labeled with value \texttt{255}.
\end{itemize}

\begin{figure*}[!ht]
\centering
\includegraphics[width=0.8\textwidth,height=0.45\textheight,keepaspectratio]{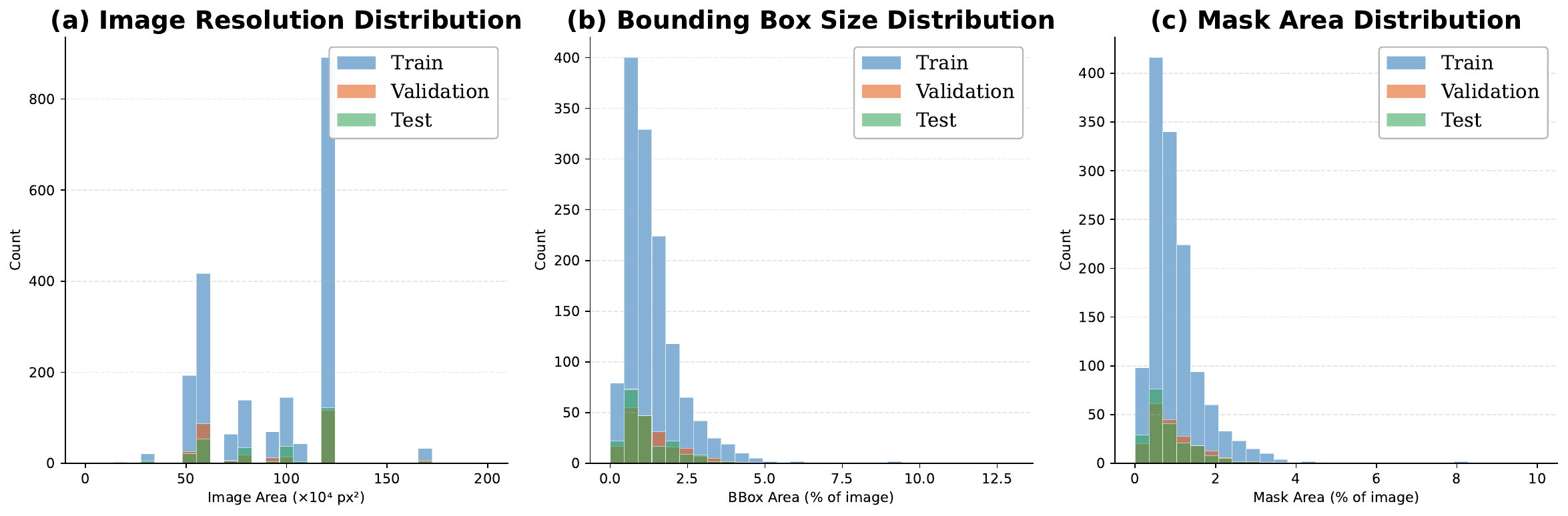}
\caption{Quantitative distribution analysis of the TAT-Det dataset: (a) distribution of ultrasound image resolutions, (b) area distribution of bounding boxes, and (c) pixel count distribution of the auxiliary segmentation masks.}
\label{fig:data_distribution}
\end{figure*}

\begin{figure}[H]
\centering
\includegraphics[width=\linewidth,height=0.45\textheight,keepaspectratio]{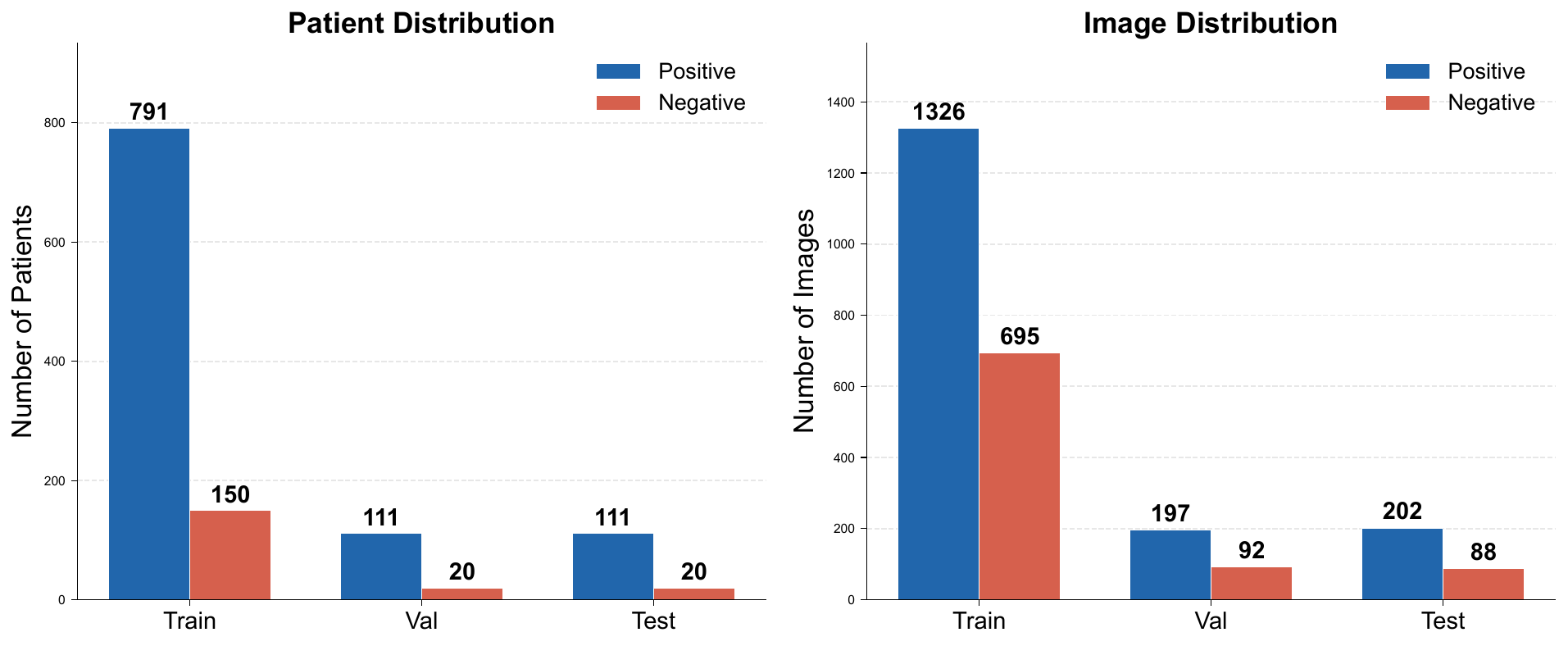}
\caption{Distribution of positive and negative samples across dataset splits.
(a) Number of unique patients per split, totaling 1{,}203 patients (941 train, 131 val, 131 test).
(b) Number of images per split, totaling 2{,}600 images (2{,}021 train, 289 val, 290 test).
Positive samples (blue) contain annotated lesions; negative samples (red) are lesion-free controls.}
\label{fig:pos_neg_distribution}
\end{figure}

\paragraph{Dataset statistics and balance}
To provide a comprehensive overview of the dataset composition, we summarize the distribution of images and key annotation statistics across the three subsets in Table~\ref{tab:dataset_stats}. Additionally, the quantitative distributions of overall image resolutions, bounding box areas, and mask pixel sizes across the dataset are visualized in Fig.~\ref{fig:data_distribution}. The distribution of positive and negative samples across dataset splits can be seen in Fig.~\ref{fig:pos_neg_distribution}.

\begin{table}[H]
\caption{Distribution of the partitioned TAT-Det dataset.}
\label{tab:dataset_stats}
\centering
\footnotesize 
\begin{tabular*}{\textwidth}{@{\extracolsep{\fill}} l c c c c }
\toprule
\textbf{Subset} & \textbf{Patients} & \textbf{Pos.} & \textbf{Neg.} & \textbf{Total Img.} \\
\midrule
\texttt{train/} & 941       & 1{,}326   & 695     & 2{,}021   \\
\texttt{val/}   & 131       & 197       & 92      & 289       \\
\texttt{test/}  & 131       & 202       & 88      & 290       \\
\midrule
\textbf{Total}  & \textbf{1{,}203} & \textbf{1{,}725} & \textbf{875} & \textbf{2{,}600} \\
\bottomrule
\end{tabular*}
\end{table}

This rigorous structured format allows researchers to easily load the dataset using standard dataloaders, enabling rapid evaluation of state-of-the-art object detection and segmentation algorithms.

\begin{figure*}[!ht]
    \centering
    \includegraphics[width=0.8\textwidth,height=0.45\textheight,keepaspectratio]{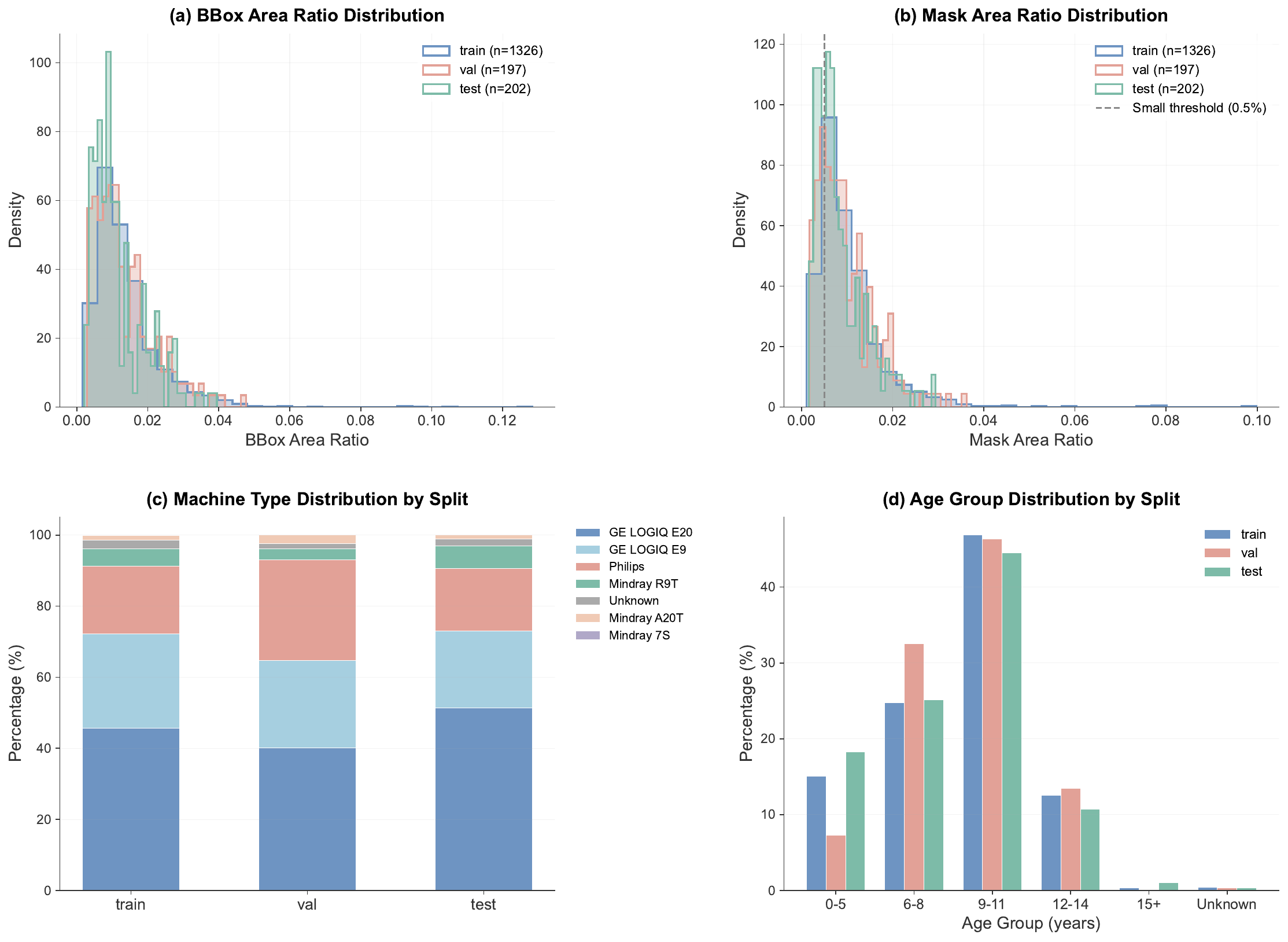}
    \caption{Dataset balance analysis across the training, validation ($n=289$), and test ($n=290$) splits. (a) BBox area ratios of positive samples ($n=1326/197/202$), showing comparable right-skewed distributions. (b) Mask area ratios; dashed line marks the small-lesion threshold ($<0.5\%$), accounting for 20.3\%, 24.9\%, and 32.7\% of positives in the three splits. (c) Machine type distribution (stacked): GE dominates (64.7--73.1\%), followed by Philips (17.6--28.4\%) and Mindray (5--9\%). (d) Age distribution (grouped bars): the 9--11 yr group is the largest (44.5--46.9\%), followed by 6--8 yr (24.8--32.5\%). Negative samples make up 34.4\%, 31.8\%, and 30.3\% of each split. No significant domain shift is observed across splits.}
    \label{fig:dataset_balance}
\end{figure*}

To evaluate the balance across the training (n=2021), validation (n=289), and test (n=290) splits of the TAT-Det dataset, we conducted a systematic analysis across four dimensions: lesion size distribution, negative sample proportion, equipment type composition, and patient age structure (see Fig.~\ref{fig:dataset_balance}). In terms of lesion size, the mean BBox area ratios across the three splits were 0.0138, 0.0136, and 0.0117, while the mean Mask area ratios were 0.0101, 0.0099, and 0.0084, respectively, with consistent medians. Small lesions (Mask area < 0.5\%) accounted for 20.3\%, 24.9\%, and 32.7\% of positive samples in the three splits. Although the test set exhibited a slightly higher proportion of small lesions, all three splits encompassed a complete range of lesion sizes from small to large, ensuring adequate training and evaluation across different lesion scales. Additionally, negative samples (lesion-free images) constituted 34.4\%, 31.8\%, and 30.3\% of the three splits, respectively, with a uniform distribution that facilitates the model's ability to learn background suppression. Regarding equipment type, GE ultrasound systems dominated all three splits (72.2\%/64.7\%/73.1\%), with Philips as the second most common (19.1\%/28.4\%/17.6\%), and Mindray series devices (R9T, A20T, 7S) collectively accounting for approximately 5\%--9\%. The distribution trends of equipment types were consistent across splits, with no excessive concentration of any single equipment type in a particular split. For patient age, the mean ages across the three splits were highly consistent (9.1/9.3/8.8 years) with comparable standard deviations (2.9/2.3/2.9). The 9--11 year age group constituted the majority in all splits (46.9\%/46.4\%/44.5\%), followed by the 6--8 year group (24.8\%/32.5\%/25.2\%), demonstrating a consistent age structure distribution. In summary, the TAT-Det dataset maintained good balance across the three splits in terms of lesion area distribution, negative sample proportion, equipment type composition, and patient age structure. No significant domain shift was detected, providing a reliable data foundation for evaluating model generalization capability.

\begin{figure}[H]
\centering
\includegraphics[width=1.0\linewidth,height=0.45\textheight,keepaspectratio]{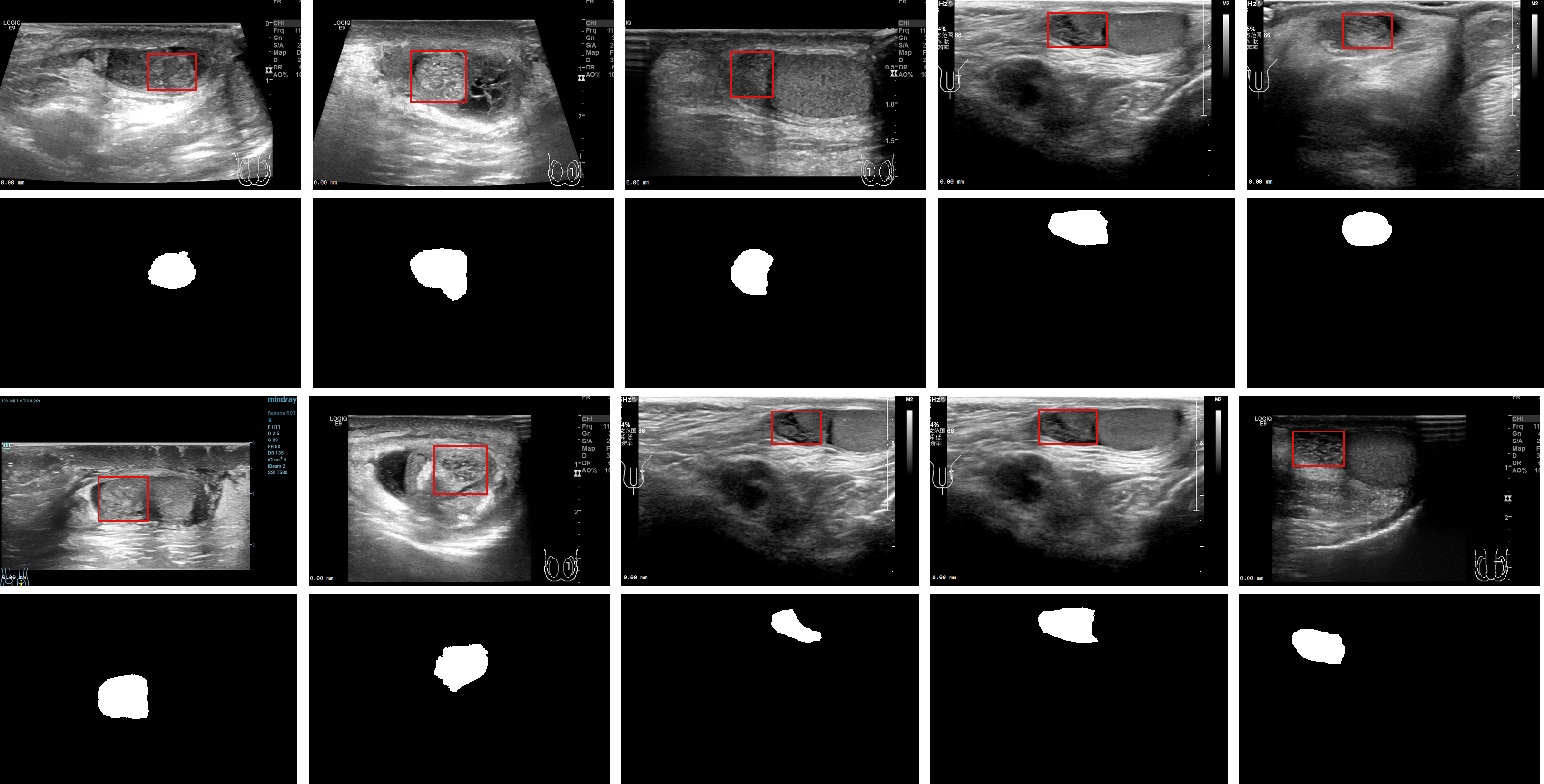}
\caption{Exemplary visualizations of annotations from the TAT-Det dataset. Rows 1 and 3 present the raw ultrasound frames overlaid with the primary ground-truth bounding boxes (red) localizing the torsed appendages. Rows 2 and 4 display the corresponding high-resolution binary segmentation masks (white) representing the clinical ground truth for the same cases directly above them.}
\label{fig:annotation_visualization}
\end{figure}

\paragraph{Visual inspection of annotations}
To visually demonstrate the high-quality and clinical precision of our expert-annotated labels, exemplar cases spanning various scanning views and lesion sizes are visualized in Fig.~\ref{fig:annotation_visualization}. The visualization is organized as a 4-row matrix to facilitate direct visual comparison: the first and third rows showcase the raw ultrasound frames overlaid with the primary ground-truth bounding boxes (red rectangles) localizing the torsed appendages, while the second and fourth rows display the corresponding binary, pixel-level segmentation masks (white on black background) for the identical clinical cases. These examples highlight the subtle contrast boundaries typical of TAT and illustrate the rigorous spatial consistency between our detection and segmentation labels.

\section{Methods}
\label{sec:methods}

\paragraph{Object detection experiments}
The baseline experiments for object detection in TAT-Det are implemented using the Ultralytics YOLO library. All experiments are conducted on a single NVIDIA GeForce RTX 4090 GPU (24 GB VRAM). Automatic mixed precision (AMP) training is enabled throughout to accelerate computation and reduce memory usage. All results are reported with deterministic training (seed=0). We evaluate seven YOLO generations spanning diverse architectures. YOLOv5~\cite{yolov5}, YOLOv8~\cite{yolov8_ultralytics}, YOLOv9~\cite{wang2024yolov9}, YOLOv10~\cite{wang2024yolov10}, YOLO11~\cite{yolo11_ultralytics}, YOLO12~\cite{tian2026yolov12}, and YOLO26~\cite{jocher2026ultralyticsyolo26unifiedrealtime} are each tested across five scales (n, s, m, l, x); YOLOv10~\cite{wang2024yolov10} includes six scales (n, s, m, b, l, x); and YOLOv9~\cite{wang2024yolov9} is evaluated with its two available variants (c, e). The input images are uniformly resized to 640$\times$640 pixels. All models are trained for 300 epochs with a batch size of 16. Pretrained weights from the MS COCO dataset~\cite{lin2014microsoft} are used for initialization. The optimizer is automatically selected by the framework (optimizer=auto), which determines the best optimizer type, initial learning rate, and momentum based on the model architecture and dataset characteristics. The final learning rate factor is set to 0.01, and a weight decay of 5e-4 is applied. A 3-epoch warmup phase is employed with an initial bias learning rate of 0.1 and a warmup momentum of 0.8. Data augmentation strategies include Mosaic augmentation (disabled in the last 10 epochs), horizontal flipping, vertical flipping, random rotation, HSV color jittering, random erasing, random translation, and Copy-Paste augmentation. The detection loss combines box regression, classification, and distribution focal loss components. Performance is evaluated on both the validation and test sets using standard COCO metrics: mAP@0.50 and mAP@0.50:0.95. All training results are systematically recorded for comparative analysis.

\paragraph{Segmentation experiments}
Baseline segmentation experiments are conducted using the APRIL-MedSeg~\cite{jiang2026april} framework. We benchmark representative CNN-based models, including UNet~\cite{ronneberger2015u}, Attention UNet~\cite{oktay2018attention}, and UNet++~\cite{zhou2018unet++}; Transformer-based models, including TransUNet~\cite{chen2021transunet} and Swin-Unet~\cite{cao2022swin}; and recent sequence-modeling architectures, including LV-UNet~\cite{jiang2024lv}, Rolling-UNet~\cite{liu2024rolling}, UKAN~\cite{li2025u}, and RWKV-UNet~\cite{jiang2025rwkv}. All experiments are conducted on a single NVIDIA GeForce RTX 4090 GPU (24 GB VRAM). Input images are resized to $256\times256$ pixels. For Swin-Unet, the input resolution is set to $224\times224$ to remain compatible with the released ImageNet pretrained weights. Unless otherwise specified, all models are trained for 200 epochs using the AdamW~\cite{loshchilov2017decoupled} optimizer (initial learning rate $1\times10^{-4}$, weight decay $1\times10^{-4}$) with a cosine learning-rate schedule~\cite{loshchilov2017sgdr} decaying to $1\times10^{-6}$. The batch size is 8, and the loss function combines cross-entropy (weight 0.4) and Dice~\cite{milletari2016v} (weight 0.6). Performance is evaluated using Dice, Intersection over Union (IoU), and the 95th percentile Hausdorff distance (HD95). All metrics are computed by comparing the predicted binary segmentation masks with the ground-truth masks, and HD95 is reported in pixel units. Background is excluded from metric computation, and the best validation checkpoint is selected for final evaluation on the test set.
\section{Analyses}
\label{sec:analyses}

To establish baseline benchmarks and to verify the reproducibility and clinical reliability of the released annotations, we conducted two sets of quantitative analyses on TAT-Det: object detection benchmarking across representative YOLO architectures and semantic segmentation benchmarking across CNN-, Transformer-, and sequence-modeling-based networks. The experimental configurations are detailed in the Methods section; the goals and basic findings of each analysis are reported here.

\subsection{Object detection benchmark}

The goal of this analysis was to evaluate whether standard deep learning detectors can reliably localize the torsed appendage on the released bounding-box annotations. Table~\ref{tab:test-results} presents the test set performance of all 33 YOLO models evaluated on the TAT-Det dataset. Among all evaluated models, YOLO11m achieves the state-of-the-art performance with the highest mAP@0.50 of 90.20\% and mAP@0.50:0.95 of 53.88\%, followed closely by YOLOv8s (89.64\%, 53.58\%) and YOLOv9c (88.06\%, 53.10\%). Notably, lightweight variants demonstrate outstanding parameter efficiency on this dataset; for instance, the legacy YOLOv5n (88.36\%, 52.42\%) and YOLOv8n (88.53\%, 52.18\%) significantly outperform many newer and much larger counterparts. Crucially, larger variants do not consistently improve performance under this dataset setting. This counterintuitive trend is evident as larger variants are consistently surpassed by their smaller counterparts: YOLOv5x (48.67\%) underperforms YOLOv5n (52.42\%), YOLOv8l (48.59\%) falls short of YOLOv8s (53.58\%), and YOLO11x (45.43\%) lags far behind YOLO11m (53.88\%). This optimization bottleneck is most pronounced in the latest architectures, which exhibit severe instability. Within the YOLO26 family, performance fluctuates: while YOLO26l obtains a robust mAP@0.50:0.95 of 52.77\%, the medium variant YOLO26m suffers a collapse, plummeting to a dataset-wide minimum of 37.25\%, and the extra-large YOLO26x (48.77\%) fails to match the smaller YOLO26s (52.38\%). Similarly, YOLO12l experiences a sharp performance drop to 45.71\% compared to YOLO12s (52.65\%). Additionally, the entire YOLOv10 generation exhibits a generalized performance deficit on this benchmark, with no single variant exceeding a mAP@0.50:0.95 of 51\%, and its lightweight YOLOv10n dropping to a low of 48.43\%. These widespread anomalies strongly indicate that the precise localization of testicular appendage torsion remains a challenge. All reported metrics are evaluated using default inference thresholds, establishing a solid baseline for future architectural designs. Example visualization results can be seen in Fig.~\ref{fig:visualization_results}.

\begin{table}[H]
\centering
\caption{Test set performance comparison of YOLO models on the TAT-Det dataset. The best results are \textbf{bolded}.}
\label{tab:test-results}
\setlength{\tabcolsep}{5pt}
\footnotesize
\renewcommand{\arraystretch}{0.95}
\begin{tabular*}{\textwidth}{@{\extracolsep{\fill}} l c c}
\toprule
\textbf{Model} & \textbf{mAP@0.50 (\%) $\uparrow$} & \textbf{mAP@0.50:0.95 (\%) $\uparrow$} \\
\midrule
YOLOv5n~\cite{yolov5} & 88.36 & 52.42 \\
YOLOv5s~\cite{yolov5} & 87.41 & 50.53 \\
YOLOv5m~\cite{yolov5} & 88.52 & 51.10 \\
YOLOv5l~\cite{yolov5} & 85.11 & 49.25 \\
YOLOv5x~\cite{yolov5} & 84.10 & 48.67 \\
\midrule
YOLOv8n~\cite{yolov8_ultralytics} & 88.53 & 52.18 \\
YOLOv8s~\cite{yolov8_ultralytics} & 89.64 & 53.58 \\
YOLOv8m~\cite{yolov8_ultralytics} & 87.65 & 50.14 \\
YOLOv8l~\cite{yolov8_ultralytics} & 82.51 & 48.59 \\
YOLOv8x~\cite{yolov8_ultralytics} & 84.82 & 49.75 \\
\midrule
YOLOv9c~\cite{wang2024yolov9} & 88.06 & 53.10 \\
YOLOv9e~\cite{wang2024yolov9} & 85.50 & 51.48 \\
\midrule
YOLOv10n~\cite{wang2024yolov10} & 84.70 & 48.43 \\
YOLOv10s~\cite{wang2024yolov10} & 88.59 & 48.59 \\
YOLOv10m~\cite{wang2024yolov10} & 84.70 & 49.20 \\
YOLOv10b~\cite{wang2024yolov10} & 84.76 & 48.90 \\
YOLOv10l~\cite{wang2024yolov10} & 85.15 & 50.84 \\
YOLOv10x~\cite{wang2024yolov10} & 85.67 & 50.18 \\
\midrule
YOLO11n~\cite{yolo11_ultralytics} & 85.16 & 49.20 \\
YOLO11s~\cite{yolo11_ultralytics} & 87.37 & 50.97 \\
YOLO11m~\cite{yolo11_ultralytics} & \textbf{90.20} & \textbf{53.88} \\
YOLO11l~\cite{yolo11_ultralytics} & 87.40 & 48.97 \\
YOLO11x~\cite{yolo11_ultralytics} & 82.00 & 45.43 \\
\midrule
YOLO12n~\cite{tian2026yolov12} & 86.51 & 51.79 \\
YOLO12s~\cite{tian2026yolov12} & 88.09 & 52.65 \\
YOLO12m~\cite{tian2026yolov12} & 85.43 & 50.53 \\
YOLO12l~\cite{tian2026yolov12} & 82.37 & 45.71 \\
YOLO12x~\cite{tian2026yolov12} & 87.54 & 52.12 \\
\midrule
YOLO26n~\cite{jocher2026ultralyticsyolo26unifiedrealtime} & 87.68 & 51.81 \\
YOLO26s~\cite{jocher2026ultralyticsyolo26unifiedrealtime} & 88.06 & 52.38 \\
YOLO26m~\cite{jocher2026ultralyticsyolo26unifiedrealtime} & 83.23 & 37.25 \\
YOLO26l~\cite{jocher2026ultralyticsyolo26unifiedrealtime} & 89.06 & 52.77 \\
YOLO26x~\cite{jocher2026ultralyticsyolo26unifiedrealtime} & 82.71 & 48.77 \\
\bottomrule
\end{tabular*}
\end{table}

\subsection{Semantic segmentation benchmark}

\begin{figure*}[!ht]
    \centering
    \includegraphics[width=\textwidth,height=0.45\textheight,keepaspectratio]{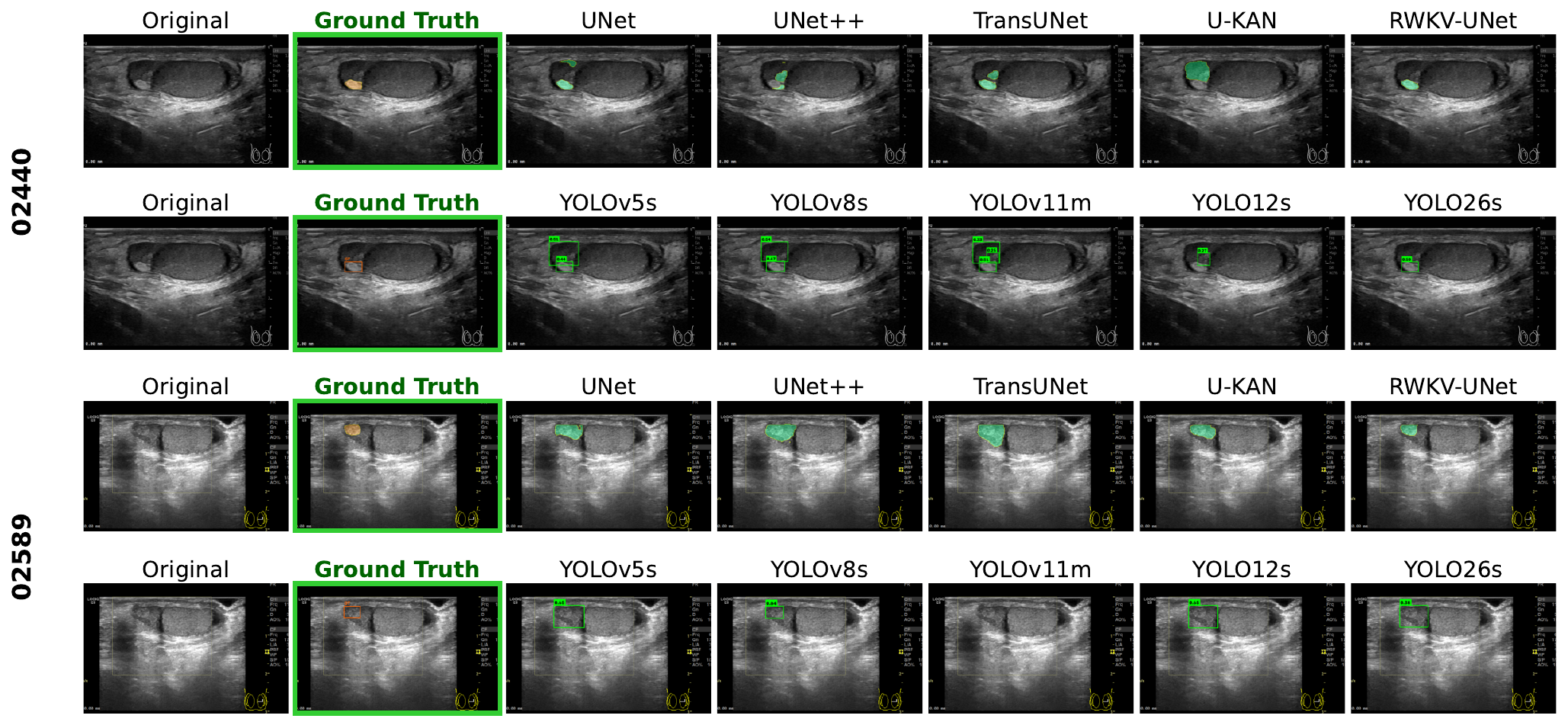}
    \caption{Qualitative comparison of representative detection and segmentation results on the TAT-Det test set.}
    \label{fig:visualization_results}
\end{figure*}

The goal of this analysis was to quantify the accuracy with which representative segmentation architectures can delineate the torsed appendage from the released pixel-level masks. Table~\ref{tab:tat_det_test} summarizes the segmentation performance of representative models on the TAT-Det test set. Conventional CNN-based models achieve comparable performance but are consistently outperformed by architectures with enhanced global context modeling. Among all evaluated methods, TransUNet substantially improves segmentation accuracy over conventional CNN baselines, while RWKV-UNet achieves the best overall performance, obtaining a Dice score of 0.8226, an IoU of 0.7861, and the lowest HD95 of 7.16. Example visualization results can be seen in Fig.~\ref{fig:visualization_results}. Overall, these baseline results demonstrate that TAT-Det supports quantitative evaluation of diverse segmentation architectures while indicating that accurate segmentation of testicular appendage torsion in ultrasound images remains challenging and leaves considerable room for future improvement.

\begin{table}[H]
\centering
\caption{Performance comparison of representative segmentation models on the TAT-Det test set. The best results are \textbf{bolded}.}
\label{tab:tat_det_test}
\footnotesize 
\begin{tabular*}{\textwidth}{@{\extracolsep{\fill}} l c c c }
\toprule
\textbf{Model} & \textbf{Dice$\uparrow$} & \textbf{IoU$\uparrow$} & \textbf{HD95$\downarrow$} \\
\midrule
UNet~\cite{ronneberger2015u}          & 76.83 & 70.45 & 11.06 \\
Att-UNet~\cite{oktay2018attention}    & 77.41 & 70.89 & 11.42 \\
UNet++~\cite{zhou2018unet++}          & 76.54 & 69.84 & 12.60 \\
TransUNet~\cite{chen2021transunet}    & 81.42 & 74.65 & 8.01  \\
Swin-Unet~\cite{cao2022swin}          & 77.16 & 69.50 & 9.06  \\
LV-UNet~\cite{jiang2024lv}            & 77.40 & 69.13 & 9.96  \\
Rolling-UNet~\cite{liu2024rolling}    & 78.61 & 71.94 & 9.42  \\
UKAN~\cite{li2025u}                   & 78.18 & 71.07 & 10.31 \\
RWKV-UNet~\cite{jiang2025rwkv}        & \textbf{82.26} & \textbf{78.61} & \textbf{7.16} \\
\bottomrule
\end{tabular*}
\end{table}

\section{Discussion}
\label{sec:discussion}

TAT-Det provides a standardized, publicly available benchmark that makes a previously under-represented pediatric emergency condition accessible to systematic computational investigation. The dataset combines a clinically relevant cohort size with expert-verified, spatially aligned detection and segmentation annotations, and its patient-level partitioning ensures that validation and test results reflect generalization to unseen patients. The balance analysis across lesion size, negative-sample proportion, equipment composition, and age structure (Fig.~\ref{fig:dataset_balance}) further indicates that no significant domain shift exists across the three splits, supporting reproducible and unbiased model evaluation.

The baseline benchmarks highlight both the utility and the difficulty of the task. Reliable localization of the torsed appendage is achievable with current detectors, as reflected by the strong performance of YOLO11m and the efficiency of lightweight variants; however, the observed instability of the latest architectures and the generalized deficit of YOLOv10 on this benchmark indicate that precise localization and delineation of small, low-contrast ultrasound lesions remain an open challenge. This is consistent with the segmentation results, where models with enhanced global context modeling (TransUNet, RWKV-UNet) outperform conventional CNN baselines, leaving considerable room for improvement. TAT-Det therefore not only supports quantitative evaluation of diverse architectures but also serves as a testbed for future algorithmic designs targeting challenging pediatric ultrasound scenarios.

\paragraph{Limitations}
TAT-Det is a retrospective single-center dataset from one pediatric hospital, and external multi-center validation is needed to assess generalizability. The current release consists predominantly of grayscale ultrasound images, with only a very small fraction containing color or spectral Doppler information. In addition, the dataset focuses on TAT detection and segmentation, while negative samples are lesion-free images rather than explicitly labeled alternative acute scrotal conditions. Additional annotation rounds involving pediatric radiologists, pediatric urologists, or multi-institutional experts can further strengthen the clinical reliability of the labels.

\section{Potential implications}
\label{sec:implications}

The dataset contains ultrasound images together with expert-annotated bounding boxes and pixel-level segmentation masks, supporting both object detection and semantic segmentation tasks. We encourage researchers to use TAT-Det as a standardized benchmark for developing, evaluating, and comparing computer-aided diagnosis methods for pediatric scrotal ultrasound.

Beyond algorithm benchmarking, TAT-Det may support a range of research and potential clinical decision-support applications. The dataset can facilitate the development of artificial intelligence-assisted systems for rapid localization and assessment of testicular appendage torsion, which may improve ultrasound-based triage and decision support in pediatric emergency care. In addition, the expert annotations and standardized data organization provide a useful resource for exploring advanced computer-vision methods, including object detection, semantic segmentation, uncertainty estimation, and foundation-model adaptation in challenging pediatric ultrasound scenarios. The dataset may also serve as an educational and research resource for reproducible benchmark development in pediatric emergency ultrasound.

\section{Data availability}
\label{sec:data_availability}

The TAT-Det dataset supporting the results of this article is released under the \textbf{Creative Commons Attribution 4.0 International License (CC BY 4.0)} and is available upon request. Owing to the sensitivity of pediatric medical imaging data, the dataset is distributed under a controlled-access model: prospective users should download the data application form from \url{https://raw.githubusercontent.com/juntaoJianggavin/TAT-Det/main/TAT-Det_Data_Application_Form.pdf}, complete it, and send the signed form to \href{mailto:juntaojiang@zju.edu.cn}{juntaojiang@zju.edu.cn}; access will be granted upon written authorization, and the dataset will be provided via a secure download link. By downloading the data, any user explicitly acknowledges, accepts, and agrees to be bound by the accompanying terms of use. Commercial use of the dataset, including commercial model training, product development, or integration into commercial services, is not permitted. All released data have been de-identified, and users are responsible for complying with all applicable ethical, institutional, and data protection requirements. Any publication resulting from the use of the TAT-Det dataset should appropriately acknowledge the dataset and cite this work.

\section{Code availability}
\label{sec:code_availability}

The object detection experiments were implemented using the open-source Ultralytics YOLO framework (\url{https://github.com/ultralytics/ultralytics}), and the semantic segmentation experiments were implemented using the open-source APRIL-MedSeg framework (\url{https://github.com/juntaoJianggavin/APRIL-MedSeg}). All benchmark configurations follow the training protocols described in the Methods section, enabling reproducible evaluation on TAT-Det.

\bibliography{tat_det}

@article{makela200719,
  title={A 19-year review of paediatric patients with acute scrotum},
  author={M{\"a}kel{\"a}, E and Lahdes-Vasama, T and Rajakorpi, H and Wikstr{\"o}m, S},
  journal={Scandinavian journal of surgery},
  volume={96},
  number={1},
  pages={62--66},
  year={2007},
  publisher={SAGE Publications Sage UK: London, England}
}

@article{kim2018clinical,
  title={Clinical features of acute scrotum in childhood and adolescence: Based on 17years experiences in primary care clinic},
  author={Kim, Jong Sung and Shin, Yu Seob and Park, Jong Kwan},
  journal={The American journal of emergency medicine},
  volume={36},
  number={7},
  pages={1302--1303},
  year={2018}
}

@book{deeg2014doppler,
  title={Doppler sonography in infancy and childhood},
  author={Deeg, Karl-Heinz and Rupprecht, Thomas and Hofbeck, Michael},
  year={2014},
  publisher={Springer}
}

@article{laimer2022multiparametric,
  title={Multiparametric ultrasound in torsion of the testicular appendages: a reliable diagnostic tool?},
  author={Laimer, Gregor and M{\"u}ller, Raphael and Radmayr, Christian and Lindner, Andrea Katharina and Lebovici, Andrei and Aigner, Friedrich},
  journal={Medical Ultrasonography},
  volume={24},
  number={1},
  pages={33--37},
  year={2022}
}

@article{varga2007acute,
  title={Acute scrotal pain in children--ten years’ experience},
  author={Varga, J and Zivkovic, D and Grebeldinger, S and Somer, D},
  journal={Urologia internationalis},
  volume={78},
  number={1},
  pages={73--77},
  year={2007},
  publisher={S. Karger AG Basel, Switzerland}
}

@article{baldisserotto2005color,
  title={Color Doppler sonography of normal and torsed testicular appendages in children},
  author={Baldisserotto, Matteo and de Souza, Jo{\~a}o Carlos Ketzer and Pertence, Ana Paula and Dora, Marcelo Dourado},
  journal={American Journal of Roentgenology},
  volume={184},
  number={4},
  pages={1287--1292},
  year={2005},
  publisher={American Roentgen Ray Society}
}

@article{sung2012sonography,
  title={Sonography of the pediatric scrotum: emphasis on the Ts—torsion, trauma, and tumors},
  author={Sung, Edward K and Setty, Bindu N and Castro-Aragon, Ilse},
  journal={American Journal of Roentgenology},
  volume={198},
  number={5},
  pages={996--1003},
  year={2012},
  publisher={American Roentgen Ray Society}
}

@article{gunther2012acute,
  title={The acute scrotum in childhood and adolescence},
  author={G{\"u}nther, Patrick and R{\"u}bben, Iris},
  journal={Deutsches {\"A}rzteblatt international},
  volume={109},
  number={25},
  pages={449},
  year={2012}
}

@article{deeg2021differential,
  title={Differential diagnosis of acute scrotum in childhood and adolescence with high-resolution duplex sonography},
  author={Deeg, Karl-Heinz},
  journal={Ultraschall in der Medizin-European Journal of Ultrasound},
  volume={42},
  number={01},
  pages={10--38},
  year={2021},
  publisher={Georg Thieme Verlag KG}
}

@article{ciftci2004clinical,
  title={Clinical predictors for differential diagnosis of acute scrotum},
  author={Ciftci, AO and {\c{S}}enocak, ME and Tanyel, F Cahit and B{\"u}y{\"u}kpamuk{\c{c}}u, N},
  journal={European journal of pediatric surgery},
  volume={14},
  number={05},
  pages={333--338},
  year={2004},
  publisher={Georg Thieme Verlag KG Stuttgart, New York{\textperiodcentered} Masson Editeur Paris}
}

@article{gomez2024bus,
  title={BUS-BRA: A breast ultrasound dataset for assessing computer-aided diagnosis systems},
  author={G{\'o}mez-Flores, Wilfrido and Gregorio-Calas, Maria Julia and Coelho de Albuquerque Pereira, Wagner},
  journal={Medical physics},
  volume={51},
  number={4},
  pages={3110--3123},
  year={2024},
  publisher={Wiley Online Library}
}

@article{yap2017automated,
  title={Automated breast ultrasound lesions detection using convolutional neural networks},
  author={Yap, Moi Hoon and Pons, Gerard and Marti, Joan and Ganau, Sergi and Sentis, Melcior and Zwiggelaar, Reyer and Davison, Adrian K and Marti, Robert},
  journal={IEEE journal of biomedical and health informatics},
  volume={22},
  number={4},
  pages={1218--1226},
  year={2017},
  publisher={IEEE}
}

@inproceedings{pedraza2015open,
  title={An open access thyroid ultrasound image database},
  author={Pedraza, Lina and Vargas, Carlos and Narv{\'a}ez, Fabi{\'a}n and Dur{\'a}n, Oscar and Mu{\~n}oz, Emma and Romero, Eduardo},
  booktitle={10th International symposium on medical information processing and analysis},
  volume={9287},
  pages={188--193},
  year={2015},
  organization={SPIE}
}

@inproceedings{gong2021multi,
  title={Multi-task learning for thyroid nodule segmentation with thyroid region prior},
  author={Gong, Haifan and Chen, Guanqi and Wang, Ranran and Xie, Xiang and Mao, Mingzhi and Yu, Yizhou and Chen, Fei and Li, Guanbin},
  booktitle={2021 IEEE 18th international symposium on biomedical imaging (ISBI)},
  pages={257--261},
  year={2021},
  organization={IEEE}
}

@article{leclerc2019deep,
  title={Deep learning for segmentation using an open large-scale dataset in 2D echocardiography},
  author={Leclerc, Sarah and Smistad, Erik and Pedrosa, Joao and {\O}stvik, Andreas and Cervenansky, Frederic and Espinosa, Florian and Espeland, Torvald and Berg, Erik Andreas Rye and Jodoin, Pierre-Marc and Grenier, Thomas and others},
  journal={IEEE transactions on medical imaging},
  volume={38},
  number={9},
  pages={2198--2210},
  year={2019},
  publisher={IEEE}
}

@article{ouyang2020video,
  title={Video-based AI for beat-to-beat assessment of cardiac function},
  author={Ouyang, David and He, Bryan and Ghorbani, Amirata and Yuan, Neal and Ebinger, Joseph and Langlotz, Curtis P and Heidenreich, Paul A and Harrington, Robert A and Liang, David H and Ashley, Euan A and others},
  journal={Nature},
  volume={580},
  number={7802},
  pages={252--256},
  year={2020},
  publisher={Nature Publishing Group UK London}
}

@article{van2018automated,
  title={Automated measurement of fetal head circumference using 2D ultrasound images},
  author={van den Heuvel, Thomas LA and de Bruijn, Dagmar and de Korte, Chris L and Ginneken, Bram van},
  journal={PloS one},
  volume={13},
  number={8},
  pages={e0200412},
  year={2018},
  publisher={Public Library of Science San Francisco, CA USA}
}

@misc{yolov5,
  title = {Ultralytics YOLOv5},
  author = {Glenn Jocher},
  year = {2020},
  howpublished = {\url{https://github.com/ultralytics/yolov5}},
  note = {Version 7.0. \url{https://doi.org/10.5281/zenodo.3908559}}
}

@misc{yolov8_ultralytics,
  author = {Glenn Jocher and Ayush Chaurasia and Jing Qiu},
  title = {Ultralytics YOLOv8},
  year = {2023},
  howpublished = {\url{https://github.com/ultralytics/ultralytics}},
  note = {Version 8.0.0}
}

@inproceedings{wang2024yolov9,
  title={YOLOv9: Learning What You Want to Learn Using Programmable Gradient Information},
  author={Wang, Chien-Yao and Yeh, I-Hau and Liao, Hong-Yuan Mark},
  booktitle={Computer Vision -- ECCV 2024},
  pages={1--21},
  year={2024},
  organization={Springer Nature Switzerland}
}

@inproceedings{wang2024yolov10,
  title={YOLOv10: Real-Time End-to-End Object Detection},
  author={Wang, Ao and Chen, Hui and Liu, Lihao and Chen, Kai and Lin, Zijia and Han, Jungong and Ding, Guiguang},
  booktitle={Advances in Neural Information Processing Systems},
  doi = {10.52202/079017-3429},
  volume={37},
  pages={107984--108011},
  year={2024}
}

@misc{yolo11_ultralytics,
  author = {Glenn Jocher and Jing Qiu},
  title = {Ultralytics YOLO11},
  year = {2024},
  howpublished = {\url{https://github.com/ultralytics/ultralytics}},
  note = {Version 11.0.0}
}

@article{tian2026yolov12,
  title={Yolov12: Attention-centric real-time object detectors},
  author={Tian, Yunjie and Ye, Qixiang and Doermann, David},
  journal={Advances in neural information processing systems},
  volume={38},
  pages={78433--78457},
  year={2026}
}

@article{langan2022scrotal,
  title={Scrotal masses},
  author={Langan, Robert C and Puente, Manuel E},
  journal={American family physician},
  volume={106},
  number={2},
  pages={184--189},
  year={2022}
}

@misc{Wada_Labelme_Image_Polygonal,
author = {Wada, Kentaro},
title = {{Labelme: Image Polygonal Annotation with Python}},
howpublished = {\url{https://github.com/wkentaro/labelme}},
note = {\url{https://doi.org/10.5281/zenodo.5711226}}
}

@misc{jocher2026ultralyticsyolo26unifiedrealtime,
  title = {Ultralytics YOLO26: Unified Real-Time End-to-End Vision Models},
  author = {Glenn Jocher and Jing Qiu and Mengyu Liu and Shuai Lyu and Fatih Cagatay Akyon and Muhammet Esat Kalfaoglu},
  year = {2026},
  eprint = {2606.03748},
  archivePrefix = {arXiv},
  primaryClass = {cs.CV},
  doi = {10.48550/arXiv.2606.03748},
  url = {https://arxiv.org/abs/2606.03748},
}

@String(ICCV= {Int. Conf. Comput. Vis.})

@String(ECCV= {Eur. Conf. Comput. Vis.})

@String(AAAI = {AAAI})

@String(ICCV  = {ICCV})

@String(ECCV  = {ECCV})

@article{al2020dataset,
  title={Dataset of breast ultrasound images},
  author={Al-Dhabyani, Walid and Gomaa, Mohammed and Khaled, Hussien and Fahmy, Aly},
  journal={Data in brief},
  volume={28},
  pages={104863},
  year={2020},
  publisher={Elsevier}
}

@inproceedings{ronneberger2015u,
  title={U-net: Convolutional networks for biomedical image segmentation},
  author={Ronneberger, Olaf and Fischer, Philipp and Brox, Thomas},
  booktitle={MICCAI},
  pages={234--241},
  year={2015},
  organization={Springer}
}

@article{oktay2018attention,
  title={Attention u-net: Learning where to look for the pancreas},
  author={Oktay, Ozan and Schlemper, Jo and Folgoc, Loic Le and others},
  journal={arXiv preprint arXiv:1804.03999},
  year={2018}
}

@article{chen2021transunet,
  title={Transunet: Transformers make strong encoders for medical image segmentation},
  author={Chen, Jieneng and Lu, Yongyi and Yu, Qihang and Luo, Xiangde and Adeli, Ehsan and Wang, Yan and Lu, Le and Yuille, Alan L and Zhou, Yuyin},
  journal={arXiv preprint arXiv:2102.04306},
  year={2021}
}

@inproceedings{cao2022swin,
  title={Swin-unet: Unet-like pure transformer for medical image segmentation},
  author={Cao, Hu and Wang, Yueyue and Chen, Joy and Jiang, Dongsheng and Zhang, Xiaopeng and Tian, Qi and Wang, Manning},
  booktitle={ECCV},
  pages={205--218},
  year={2022},
  organization={Springer}
}

@article{jiang2025rwkv,
  title={Rwkv-unet: Improving unet with long-range cooperation for effective medical image segmentation},
  author={Jiang, Juntao and Zhang, Jiangning and Liu, Weixuan and Gao, Muxuan and Hu, Xiaobin and Xue, Zhucun and Liu, Yong and Yan, Shuicheng},
  journal={arXiv preprint arXiv:2501.08458},
  year={2025}
}

@inproceedings{kirillov2023segment,
  title={Segment anything},
  author={Kirillov, Alexander and Mintun, Eric and Ravi, Nikhila and others},
  booktitle={ICCV},
  pages={4015--4026},
  year={2023}
}

@inproceedings{jiang2024lv,
  title={LV-UNet: a lightweight and vanilla model for medical image segmentation},
  author={Jiang, Juntao and Wang, Mengmeng and Tian, Huizhong and Cheng, Lingbo and Liu, Yong},
  booktitle={2024 IEEE International Conference on Bioinformatics and Biomedicine (BIBM)},
  pages={4240--4246},
  year={2024},
  organization={IEEE}
}

@inproceedings{liu2024rolling,
  title={Rolling-unet: Revitalizing mlp’s ability to efficiently extract long-distance dependencies for medical image segmentation},
  author={Liu, Yutong and Zhu, Haijiang and Liu, Mengting and Yu, Huaiyuan and Chen, Zihan and Gao, Jie},
  booktitle={Proceedings of the AAAI conference on artificial intelligence},
  volume={38},
  number={4},
  pages={3819--3827},
  year={2024}
}

@inproceedings{zhou2018unet++,
  title={Unet++: A nested u-net architecture for medical image segmentation},
  author={Zhou, Zongwei and Rahman Siddiquee, Md Mahfuzur and Tajbakhsh, Nima and Liang, Jianming},
  booktitle={International workshop on deep learning in medical image analysis},
  pages={3--11},
  year={2018},
  organization={Springer}
}

@inproceedings{li2025u,
  title={U-kan makes strong backbone for medical image segmentation and generation},
  author={Li, Chenxin and Liu, Xinyu and Li, Wuyang and Wang, Cheng and Liu, Hengyu and Liu, Yifan and Chen, Zhen and Yuan, Yixuan},
  booktitle={Proceedings of the AAAI conference on artificial intelligence},
  volume={39},
  number={5},
  pages={4652--4660},
  year={2025}
}

@article{jiang2026april,
  title={APRIL-MedSeg: A Modular Medical Image Segmentation Toolbox Embracing Modern Paradigms},
  author={Jiang, Juntao and Bai, Jinsheng and Fan, Linxuan and Bi, Yali and Zhang, Jiangning and Liu, Yong},
  journal={arXiv preprint arXiv:2606.30577},
  year={2026}
}

@inproceedings{milletari2016v,
  title={V-net: Fully convolutional neural networks for volumetric medical image segmentation},
  author={Milletari, Fausto and Navab, Nassir and Ahmadi, Seyed-Ahmad},
  booktitle={2016 fourth international conference on 3D vision (3DV)},
  pages={565--571},
  year={2016},
  organization={Ieee}
}

@article{loshchilov2017decoupled,
  title={Decoupled weight decay regularization},
  author={Loshchilov, Ilya and Hutter, Frank},
  journal={arXiv preprint arXiv:1711.05101},
  year={2017}
}

@inproceedings{loshchilov2017sgdr,
  title={Sgdr: Stochastic gradient descent with warm restarts},
  author={Loshchilov, Ilya and Hutter, Frank},
  booktitle={International conference on learning representations},
  year={2017}
}

@inproceedings{lin2014microsoft,
  title={Microsoft coco: Common objects in context},
  author={Lin, Tsung-Yi and Maire, Michael and Belongie, Serge and Hays, James and Perona, Pietro and Ramanan, Deva and Doll{\'a}r, Piotr and Zitnick, C Lawrence},
  booktitle={European conference on computer vision},
  pages={740--755},
  year={2014},
  organization={Springer}
}

\end{document}